\documentclass{article}

\usepackage[final]{nips_2016}

\usepackage[utf8]{inputenc} 
\usepackage[T1]{fontenc}    
\usepackage{hyperref}       
\usepackage{url}            
\usepackage{booktabs}       
\usepackage{amsfonts}       
\usepackage{amsmath}
\usepackage{nicefrac}       
\usepackage{microtype}      
\usepackage{graphicx}
\usepackage{subfigure}
\usepackage[ruled,vlined,linesnumbered,noresetcount]{algorithm2e}
\usepackage{varwidth}
\usepackage[noend]{algpseudocode}
\pdfoutput=1
\graphicspath {{figures/}}
\begin{document}

\title{Speeding up Convolutional Neural Networks By Exploiting the Sparsity of Rectifier Units}

\author{
  Shaohuai Shi\\
  Department of Computer Science\\
  Hong Kong Baptist University\\
  Kowloon, Hong Kong\\
  \texttt{csshshi@comp.hkbu.edu.hk} \\
  \and
  \textbf{Xiaowen Chu}\\
  Department of Computer Science\\
  Hong Kong Baptist University\\
  Kowloon, Hong Kong\\
  \texttt{chxw@comp.hkbu.edu.hk} \\
  }

\maketitle

\begin{abstract}
Rectifier neuron units (ReLUs) have been widely used in deep convolutional networks. An ReLU converts negative values to zeros, and does not change positive values, which leads to a high sparsity of neurons. In this work, we first examine the sparsity of the outputs of ReLUs in some popular deep convolutional architectures. And then we use the sparsity property of ReLUs to accelerate the calculation of convolution by skipping calculations of zero-valued neurons. The proposed sparse convolution algorithm achieves some speedup improvements on CPUs compared to the traditional matrix-matrix multiplication algorithm for convolution when the sparsity is not less than 0.9.
\end{abstract}

\section{Introduction} \label{introduction}
Deep neural networks (DNNs) have been succefully used in many AI applications \cite{lecun2015deep}. DNNs were proposed in the early 1982 \cite{fukushima1982neocognitron}. With the help of the increasing computional ability of general-purpose processing units like GPUs, DNNs enjoy a fast expansion in recent years. One of the important reasons why DNNs have a high ability to fitting data is that its sparse representation of an input space \cite{willmore2001characterizing}. The rectifier neuron units (i.e., ReLUs) contribute a lot to the success of deep neural networks \cite{glorot2011deep,nair2010rectified,maas2013rectifier,lecun2015deep}. ReLUs can not only solve the vanishing gradient problem, but they also produce a sparse representation which leads to mathematical advantages of the network \cite{nair2010rectified}. Given the sparse structure of DNNs, the networks have many zero operands. These zero operands can be neglected to reduce the magnitude of calculation since multiply and addition with zero are unmeaningful. Albericio et al. \cite{albericio2016cnvlutin} have shown that the inputs of convolutional layers are on average 44\% zeros based on recently popular deep learning architectures, and they propose a DNN accelerator named \textit{Cnvlutin} to dymanically eliminate the most inefectural multiplications of zeros on the simulator of \textit{DaDianNao} which is a state-of-the-art accelerator proposed by Chen et al. \cite{chen2014dadiannao}.

In this work, we also exploit the property of ReLUs to accelate the computation of DNNs, but in two different ways compared with \cite{albericio2016cnvlutin}: First, we observe the sparsity of output values with ReLU activivation functions in the existing popular deep architectures during the training process but not limited to the inferencing phase. Second, we design the algorithm that eliminates the unused zero operands on-the-fly on traditional processors (i.e., CPUs) without changing the architecture of network. Our proposed convolution algorithm for the sparsity that is not less than 0.9 achieves performance improvements on convolutions of LeNet \cite{lecun1998gradient}, AlexNet \cite{krizhevsky2012imagenet} and GoogLeNet \cite{szegedy2015going} on CPUs compared to the traditional convolution algorithm.

\section{Related Work}
\label{s:relatedwork}
The sparsity of parameters have been widely exploited to compress and accelerate DNNs \cite{han2015learning,changpinyo2017power,han2015deep,liu2015sparse,guo2016dynamic,wen2016learning}. DNNs with large size of parameters have been empirically proved that the parameters exist heave redundancy which can be removed by sparse decompositions \cite{liu2015sparse}, sparse learning \cite{wen2016learning}, continual network maintenance \cite{guo2016dynamic} and guided pruning \cite{park2017sparse} etc. This kind of methods focus on constructing the high sparsity of parameters and pruning unused connections to achieve the network compression and acceleration. It needs to carefully construct on the sparsity of parameters since, otherwise it may lead to the accuracy decrease of original networks. Though this work does not attempt to construct the sparsity of deep networks, the fact of high speedup in the sparse networks inspires us to consider the efficient convolution algorithms by skipping the calculation of zero. In fact, the ReLU activation functions set all negative values to zero, which means the input of the next layer has large propotion of zero values. This property gives oppotunities for energy saving \cite{chen2016eyeriss,reagen2016minerva} and MAC reducing \cite{albericio2016cnvlutin} by skipping the insignificiant calculation of zero. Reagen et al. \cite{reagen2016minerva} state that operations involving large number of zero or near zero values can save power both from avoided multiplications and SRAM accesses. Albericio et al. \cite{albericio2016cnvlutin} show that there are on average 50\% zero of convolutional inputs in seven state-of-the-art deep convolutional networks, and they design an efficient convolution accelerator by skipping the calculation of zeros. In terms of the high sparsity of input of convolution, few work was proposed to improve the efficiency of convolution on traditional computational units like CPUs and GPUs. In this work, we first demostrate the sparsity of deep neural networks whoes activation functions are ReLUs during the process of training. Second, instead of considering the particular hardware for convolution, we propose an efficient convolution algorithm that can eliminates the unused zero operands on-the-fly on CPUs without changing the archetecture of network.

\section{Sparsity of ReLUs}
\label{s:analysis}
\begin{figure}[!ht]
  \centering
    \includegraphics[width=0.9\linewidth]{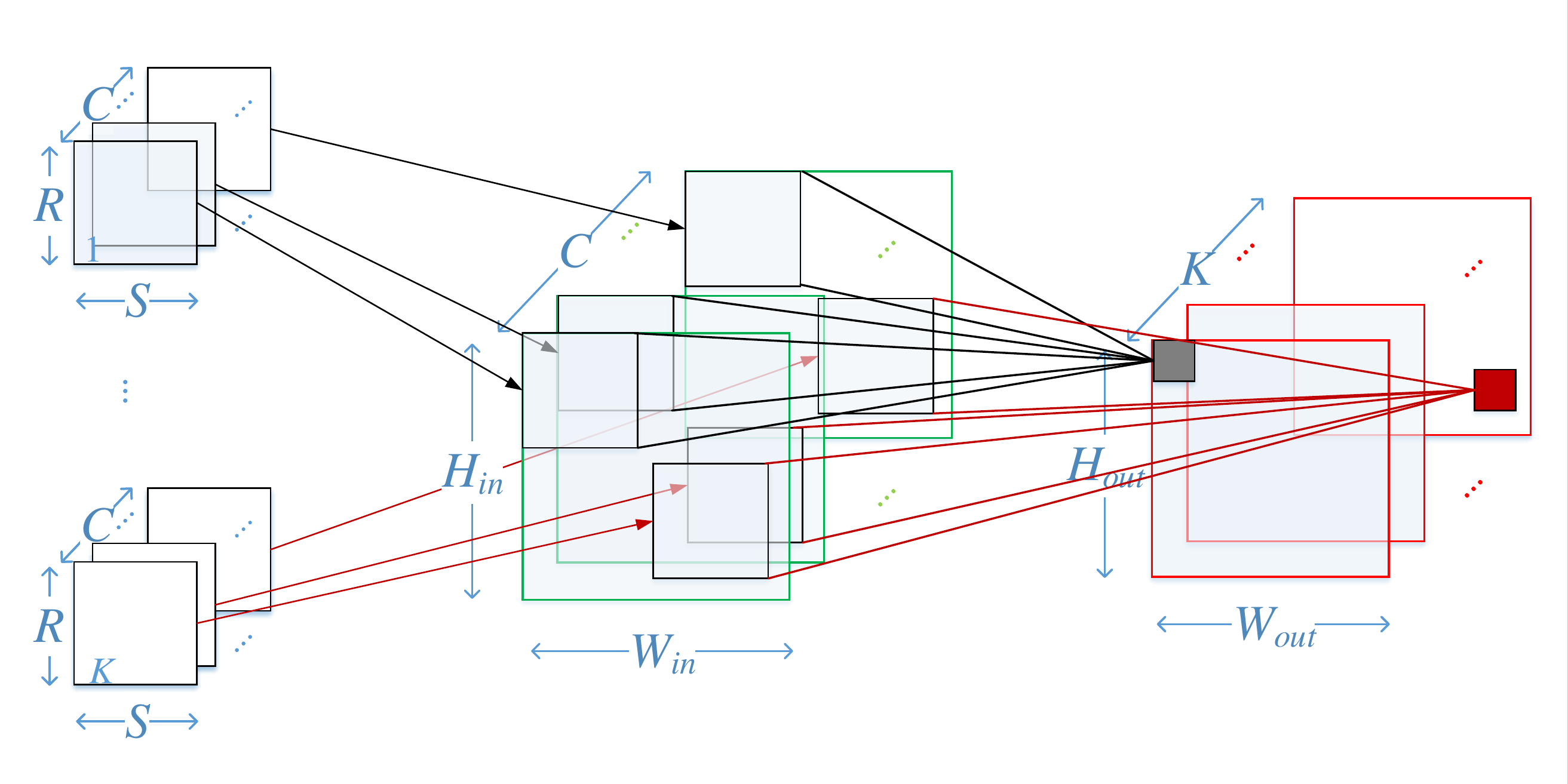}
\caption{Convolution in CNNs}
\label{fig:convolution}
\end{figure}

In this section, we want to demostrate the propobility of zero outputs of a convolutional nueral network. ReLU is an activation function $f$ with form of: $f(x)=max(0, x)$. Given a convolutional layer with $C$ channels of input feature maps, $K$ channels of output feature maps, the number of filters should be $C\times K$. Some notations are shown in Table \ref{introduction}, and we denote the convolution as the operation of high dimension tensors as shown in Fig. \ref{fig:convolution}. The input $\mathcal{I}$ is a 3-mode tensor with size of $C\times H_{in} \times W_{in}$, the kernel (weight) $\mathcal{W}$ is a 4-mode tensor with size $K\times C\times R\times S$, and the output $\mathcal{O}$ is a 3-mode tensor with size $K\times H_{out}\times W_{out}$, where $H_{out}=\frac{(H_{in}+2\times P-R+1)}{T}$ and $W_{out}=\frac{(W_{in}+2\times P-S+1)}{T}$.

\begin{table}[!ht]
\centering
\caption{Parameters of Convolution}
\label{table:parameters}
\begin{tabular}{|l|l|}
\hline
Parameter  &  Meaning \\\cline{1-2}
\hline
\hline
$C$ & \# of input feature maps \\\cline{1-2}
$H_{in}$ & Height of input image \\\cline{1-2}
$W_{in}$ & Width of input image \\\cline{1-2}
$K$ & \# of output feature maps \\\cline{1-2}
$R$ & Height of filter kernel \\\cline{1-2}
$S$ & Width of filter kernel \\\cline{1-2}
$H_{out}$ & Height of output image \\\cline{1-2}
$W_{out}$ & Width of output image \\\cline{1-2}
$P$ & Padding of input image \\\cline{1-2}
$T$ & Stride of convolution \\\cline{1-2}
\end{tabular}
\end{table}

The output value at row $i$ and column $j$ of $k$th output channel is calculated as follows:
\begin{equation}
\label{equation:conv}
\mathcal{O}_{k,i,j}=\sum_{c=0}^{C-1}\sum_{r=0}^{R-1}\sum_{s=0}^{S-1}\mathcal{W}_{k,c,r,s}\mathcal{I}_{c,i+r,j+s}
\end{equation}

In general, the convolutional layer is followed by an activation layer (e.g., ReLU), whose dimension is same with the output of convolutional layer. Let $\mathcal{Z}$ denote the output of an ReLU layer, which is generally the input of the next convolutional layer in DNNs \cite{szegedy2015going,he2016deep}. We have:
\begin{equation}
\label{equation:relu}
\mathcal{Z}_{k,i,j}=max(0, \mathcal{O}_{k,i,j})
\end{equation}

\begin{figure}[!htbp]
  \centering
  \subfigure[LeNet]
  {
    \includegraphics[width=0.45\linewidth]{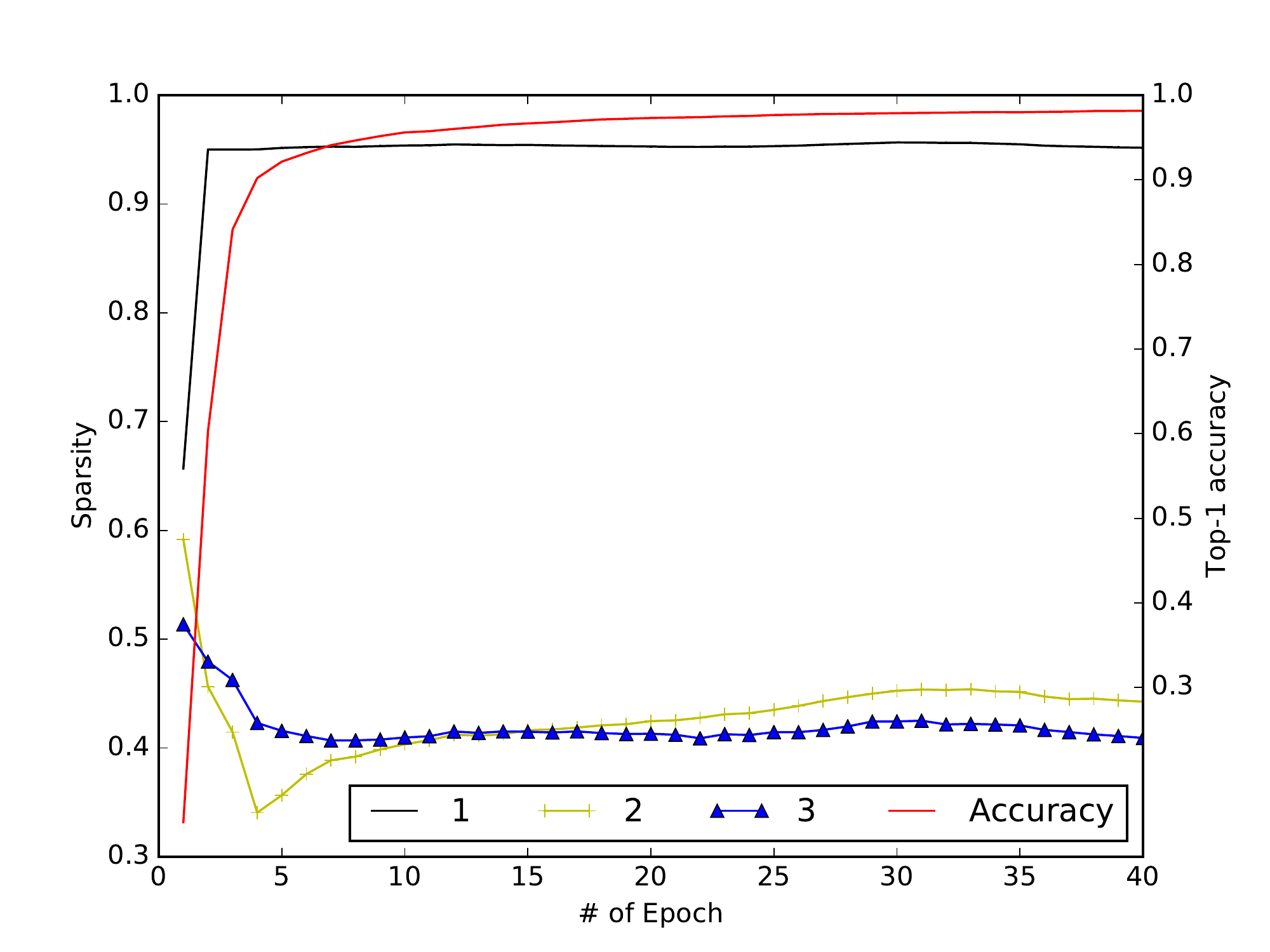}
    \label{fig:sparle}
  }
  \subfigure[AlexNet on Cifar10]
  {
    \includegraphics[width=0.45\linewidth]{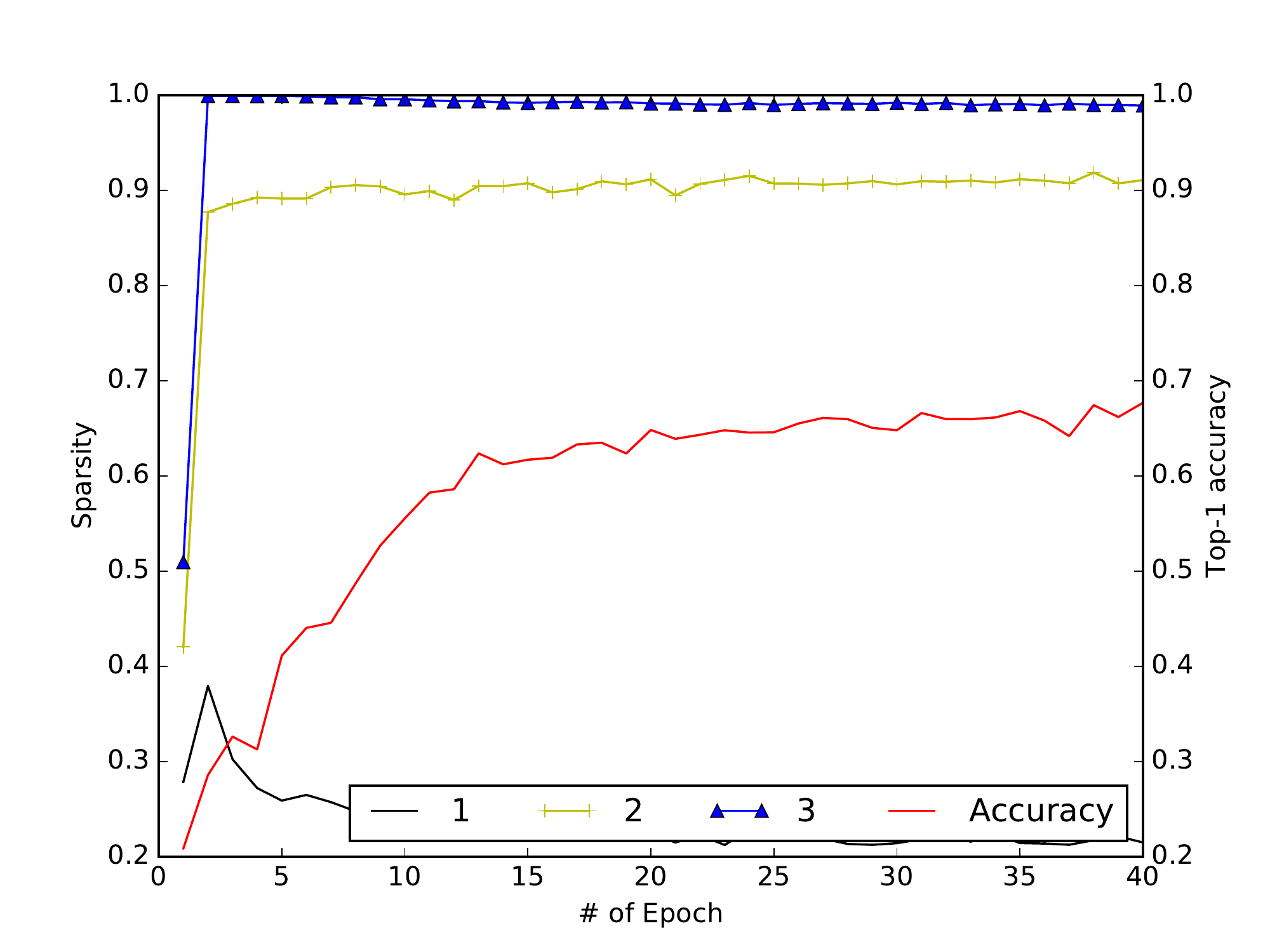}
    \label{fig:sparac}
  }
  \subfigure[AlexNet on ImageNet]
  {
    \includegraphics[width=0.45\linewidth]{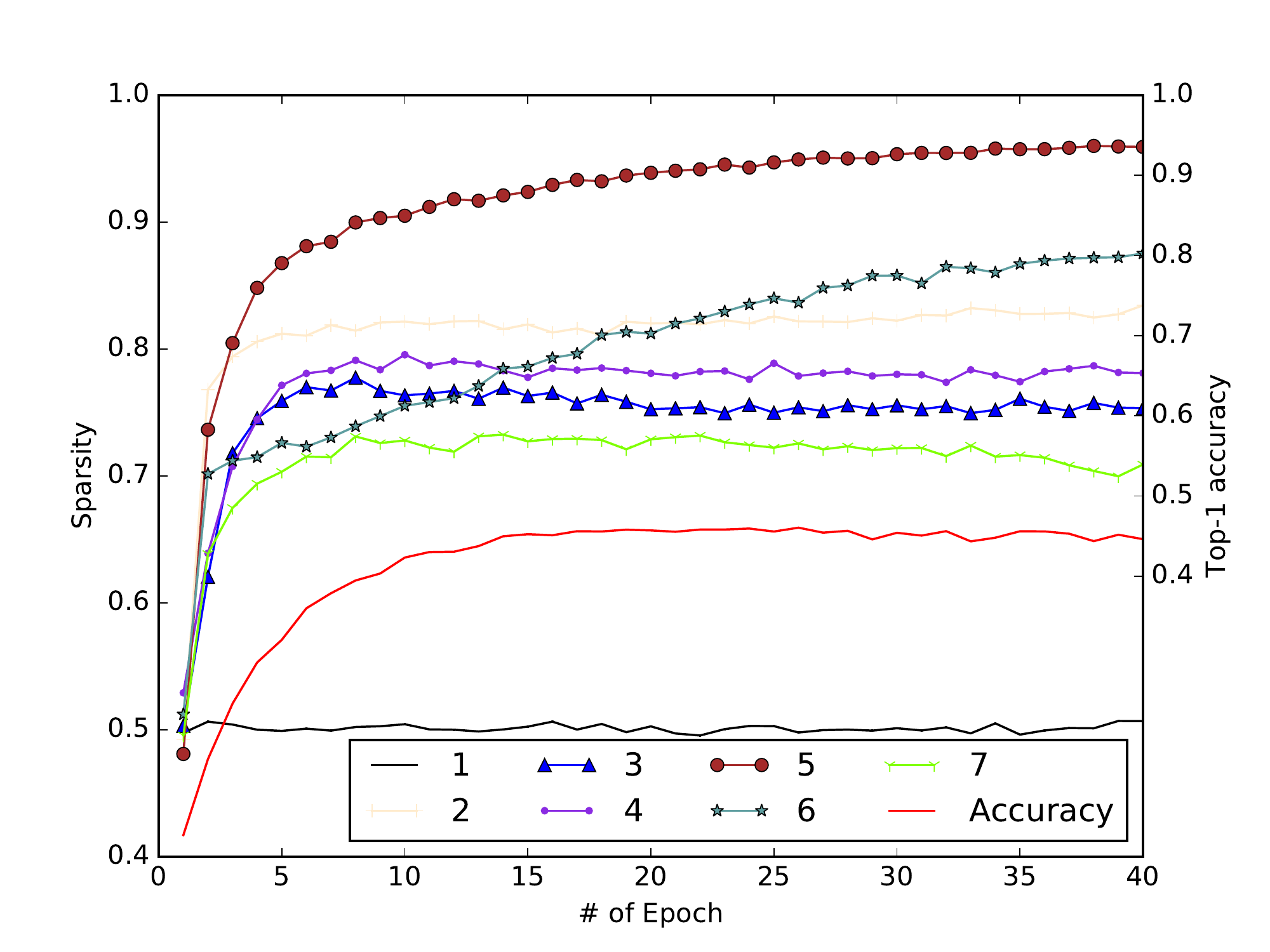}
    \label{fig:sparai}
  }
  \subfigure[GoogLeNet]
  {
    \includegraphics[width=0.45\linewidth]{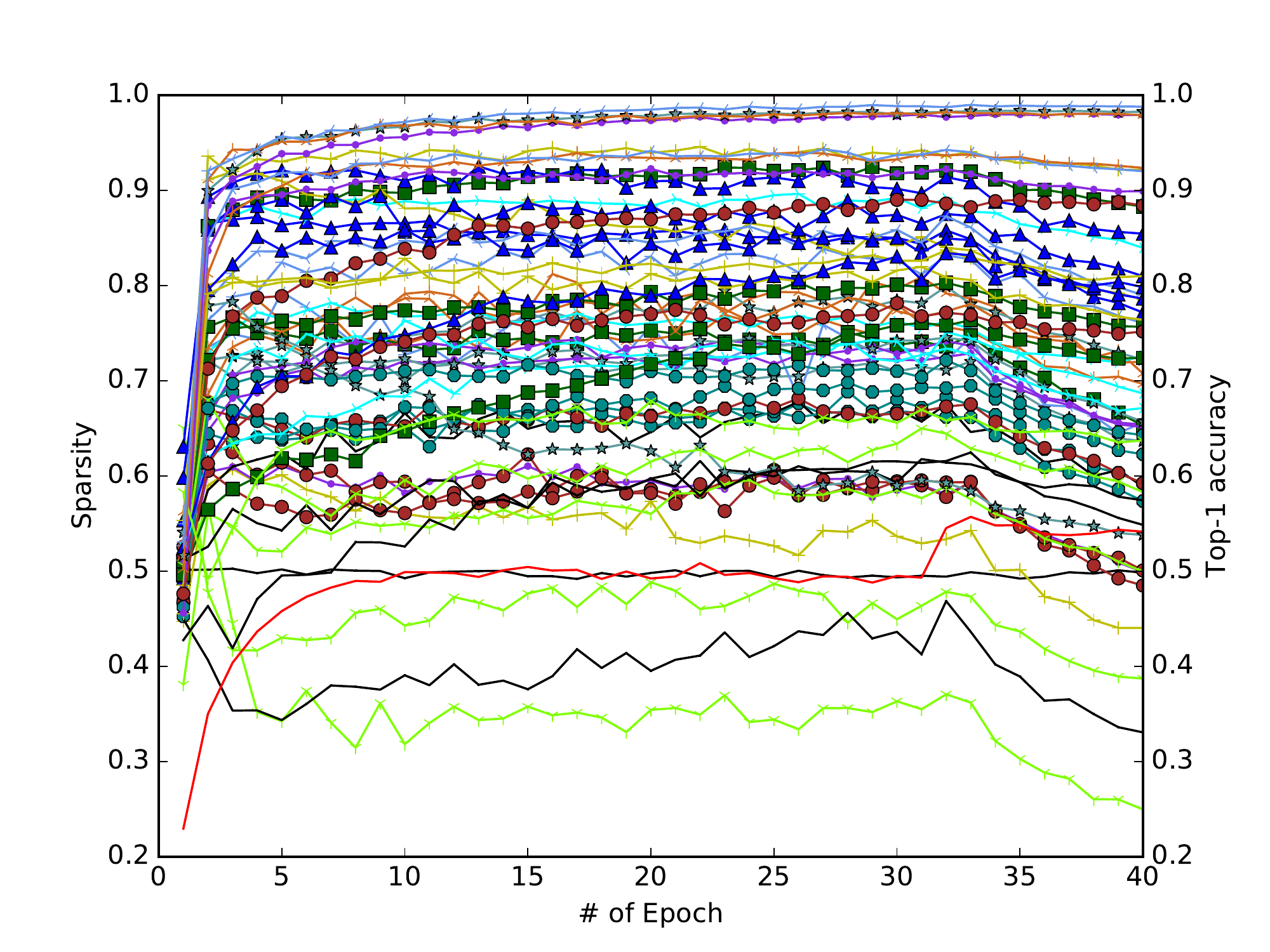}
    \label{fig:spargooglenet}
  }
\caption{The sparsity of each output layer on popular deep neural networks. The number in the legend represents the $n$th ReLU layer in the deep nerual networks. The red line without markers indicates the validation accuracy.}
\label{fig:sparcnn}
\end{figure}

In the training progress in practical, the sparsity of output of ReLUs is around 50\% to 95\%. We test the output sparsity of ReLUs in the original DNNs including LeNet \cite{lecun1998gradient} for MNIST dataset, AlexNet \cite{krizhevsky2012imagenet} for Cifar10 dataset, AlexNet \cite{krizhevsky2012imagenet} for ImageNet and GoogLeNet \cite{szegedy2015going} for ImageNet. The training process runs 40 epochs for each network. At the end of each epoch, both the sparsities of all the ReLU Layers and validation accuracy are recorded. The final results are shown in Fig. \ref{fig:sparcnn}. It is noted that many of ReLU layers generate the sparsity more than 0.7, and some of layers have up to the sparsity of about 0.95. In general, the output of an ReLU layer is the input the convolutional layer. In other words, the input of the convolution layer has up to more than 70\% zero values in most of cases. So we have the evidence to propose the sparse convolution algorithm by skipping the multiplication of zero-valued neurons to accelerate the computation of convolution.

\section{Sparse Convolution}
A direct convolution operation in CNNs takes $H_{out}\times W_{out} \times R \times S \times K \times C$ floating-point multiplication and addition calculations (MACs in \textit{FLOP}). And it needs $H_{out}\times W_{out}\times K$ memory write operations (MWOs) to store outputs. In order to exploit the advantage of SIMD techniques of processors (CPUs or GPUs), the calculation of convolutional layer is often mapped to matrix multiplication \cite{chetlur2014cudnn,sze2017efficient}. So that it can make use of highly optimized BLAS libraries like Intel MKL on CPUs and cuBLAS on GPUs. We first have a look at the matrix multiplication algorithm to solve the convolution, and analysis its computation performance. After that, we derive our accelerated algorithm by skipping zero operations.

\begin{figure}[!ht]
  \centering
    \includegraphics[width=0.9\linewidth]{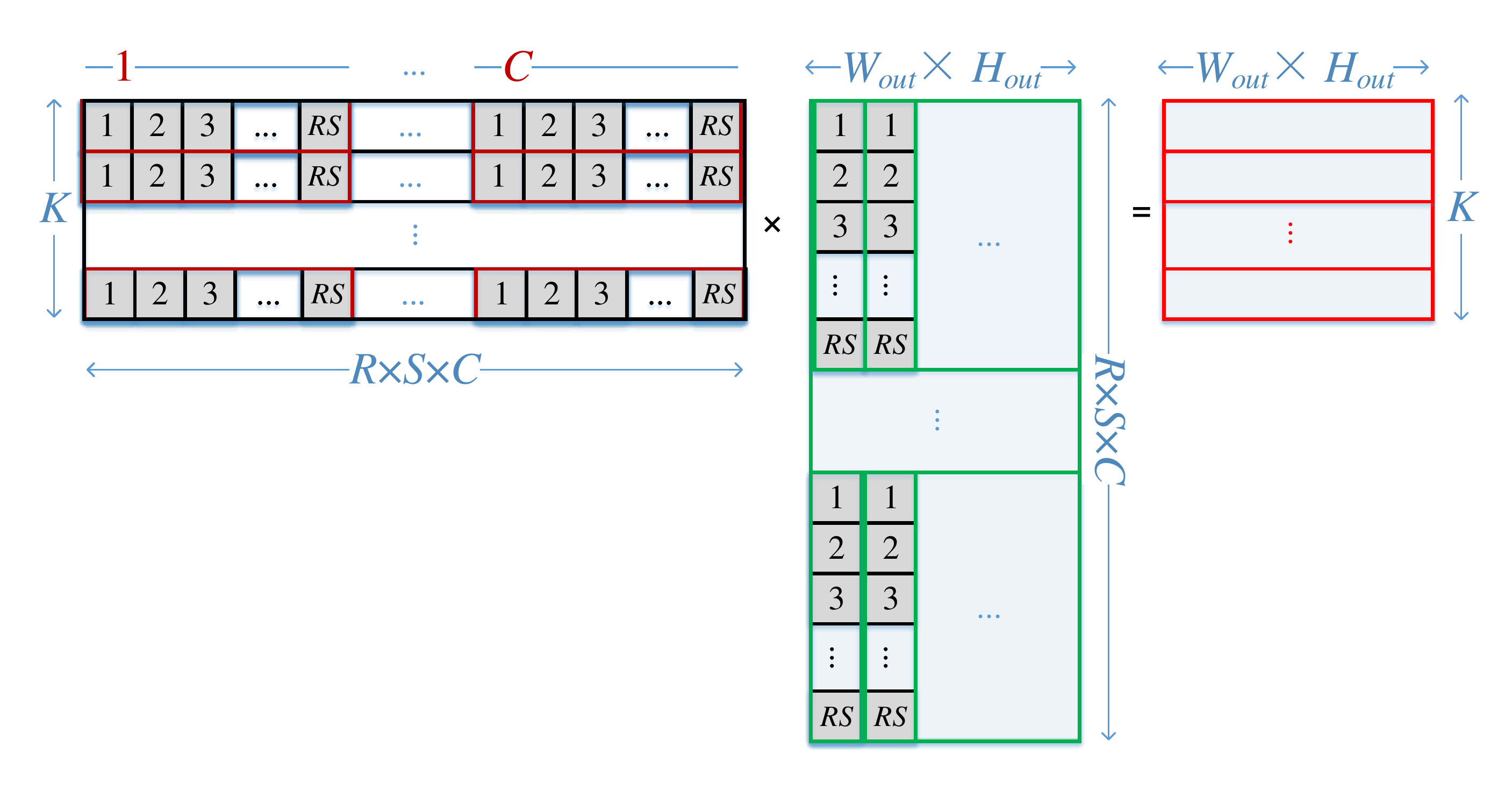}
\caption{Matrix multiplication for convolution.}
\label{fig:gemmconv}
\end{figure}
\textbf{Matrix Multiplication for Convolution.} Fig. \ref{fig:gemmconv} shows the matrix multiplication: $M_K\times M_I=M_O$ for convolution, where $M_K \in \mathbb{R}^{K\times RSC}$ is the matrix of kernels, $M_I \in \mathbb{R}^{RSC\times H_{out}W_{out}}$ is the unrolled matrix of input neurons and $M_O \in \mathbb{R}^{K\times H_{out}W_{out}}$ is the output of convolution. Note that the MACs and MWOs required by this algorithm is the same with the direct convolution method. It is easy to use SIMD and parallel techniques to accelerate matrix multiplication algorithm. 

\textbf{Inverse Sparse Convolution.} As we show in the Section \ref{s:analysis}, there are not less than 50\% zeros after the activation of ReLU, whose output is generally the input of the next convolutional layer. We design an algorithm to accelerate the convolution operation by skipping the calculations involved with zeros. In other words, the zero values are excluded to calculating the convolutional output, which reduces up to 50\% multiplication and addition operations. One may think that this skipping-zero calculations can be mapped to dense-sparse matrix multiplication. However, the dense-sparse algorithm is mostly slower than the dense-dense case since the memory access in the dense case can be continuous \cite{liu2015sparse}. So it is chanllenging to speedup the convolution with sparse input. 

Instead of scanning each element of outputs, we scan each non-zero input, and calculate the multiplication with kernels to generate the temporary results of outputs. We have two considerations of the proposed algorithm: on one hand, the MACs can be reduce to $\rho\times H_{out}\times W_{out} \times R \times S \times K \times C$, given a sparsity: $1-\rho$ of input $\mathcal{I}$, where $\rho=\frac{\#OfNonZeros}{C\times H_{in}\times W_{in}}$. On the other hand, the number of memory write operations of the ouput results should not be too large. For each non-zero $\mathcal{I}_{c,i,j}$ in inputs, calculate the product: $\mathcal{I}_{c,i,j}\times \mathcal{W}_{k,c,i-r,j-s}$ for an output: $\mathcal{O}_{k,i+r,j+s}$. 

With the two considerations above, we propose the inverse sparse convolution (ISC) algorithm by three steps: First, we skip all the zero elements of the input data, and store the non-zero values in a vector with their column and row information. Second, the kernel matrix is stored as column-major matrix such that for each non-zero element ($\mathcal{I}_{c,i,j}$) of inputs, a continious memory that stores kernels (say 4 floats of weights: $\mathcal{W}_{k,c,i-r,j-s},\mathcal{W}_{k+1,c,i-r,j-s},\mathcal{W}_{k+2,c,i-r,j-s}$ and $\mathcal{W}_{k+3,c,i-r,j-s}$) can be fetched and multiplied by $\mathcal{I}_{c,i,j}$ at one time with AVX or SSE techniques. Third, transpose temporary results from the second step to generate outputs: $\mathcal{O}$. 

\section{Experiments}
The baseline of convolution algorithm on CPUs is the GEMM version from Caffe \cite{jia2014caffe} with MKL library, which is faster than other deep learning frameworks with single core of CPUs \cite{shi2016benchmarking}. There exist fast convolution algorithms like Winograd and FFT based algorithms \cite{lavin2016fast}, but they need prerequisite for usage. The Wingrad based algorithm is used for the kernel whose size is 3$\times$3, while the FFT based algorithm requires the value of \textit{N} should be equal to the power of 2. We run the speed measurements with convolution layers whose sparsity of inputs is high enough (i.e., $\geq$ 0.9) on four popular deep networks (i.e., LeNet, AlexNet for Cifar10, AlexNet \cite{krizhevsky2012imagenet} for ImageNet, GoogLeNet \cite{szegedy2015going}). The experimental convolution layers are summarized in Table \ref{table:cnnsum}.

\begin{table}[!ht]
\centering
\caption{Convolution layers on different deep networks. The sparsity indicates the ratio of zero-valued input over the total number of input in that layer. AlexNetC and AlexNetI represent AlexNet for Cifar10 and ImageNet respectively.}
\label{table:cnnsum}
\begin{tabular}{|c|c|c|c|c|c|c|c|c|}
\hline
Network &  Layer & $C$ & $H_{in}\times W_{in}$ & $K$ & $R\times S$ & $P$ & $T$ & Sparsity\\\cline{1-9}
\hline
\hline
LeNet & Conv2 & 20 & 11$\times$11 & 64 & 5$\times$5 & 1 & 2 &0.95\\\cline{1-9}
AlexNetC & Conv3 & 32 & 6$\times$6 & 64 & 5$\times$5 & 2 & 1 &0.9\\\cline{1-9}
AlexNetI & Conv4 & 384 & 5$\times$5& 384 & 3$\times$3 & 1 & 1 & 0.9\\\cline{1-9}
GoogLeNet & Inception4a.1 & 480& 14$\times$14& 192 & 1$\times$1 & 0 & 1 & 0.9\\\cline{1-9}
GoogLeNet & Inception4a.2 & 192& 14$\times$14& 96 & 1$\times$1 & 0 & 1 & 0.9\\\cline{1-9}
GoogLeNet & Inception4e.3 & 160& 14$\times$14& 320 & 3$\times$3 & 1 & 1 & 0.9\\\cline{1-9}
GoogLeNet & Inception5a.1 & 832 & 7$\times$7& 256 & 1$\times$1 & 0 & 1 & 0.95\\\cline{1-9}
GoogLeNet & Inception5a.2 & 256&  7$\times$7& 160 & 1$\times$1 & 0 & 1 & 0.9\\\cline{1-9}
GoogLeNet & Inception5b.3 & 192 & 7$\times$7& 384 & 3$\times$3 & 1 & 1 & 0.95 \\\cline{1-9}
GoogLeNet & Inception5b.5 & 48 & 7$\times$7& 128 & 5$\times$5 & 2 & 1 & 0.95 \\\cline{1-9}

\end{tabular}
\end{table}

We measure the speed of compared algorithms on the Interl CPU: E5-2630v4 at the core frequency of 2.20GHz with 128 GB memory. The experimental results are shown in Table \ref{table:cnnresults}.
\begin{table}[!ht]
\centering
\caption{Experimental results (Time in msec).}
\label{table:cnnresults}
\begin{tabular}{|c|c|c|c|c|}
\hline
	Layer&	 Sparsity &GEMM & ISC	&Speedup\\\cline{1-5}
\hline
\hline
LeNet-Conv2&    0.95    &0.854  &0.123  &6.96X \\\cline{1-5}
AlexNetC-Conv3& 0.9     &0.329  &0.122  &2.69X \\\cline{1-5}
AlexNetI-Conv4& 0.9     &4.874  &1.660  &2.94X \\\cline{1-5}
GoogLeNet-Inception4a.1&        0.9     &1.795  &1.292  &1.39X \\\cline{1-5}
GoogLeNet-Inception4a.2&        0.9     &0.943  &0.558  &1.69X \\\cline{1-5}
GoogLeNet-Inception4e.3&        0.9     &7.602  &4.782  &1.59X \\\cline{1-5}
GoogLeNet-Inception5a.1&        0.95    &1.912  &0.645  &2.97X \\\cline{1-5}
GoogLeNet-Inception5a.2&        0.9     &0.742  &0.258  &2.87X \\\cline{1-5}
GoogLeNet-Inception5b.3&        0.95    &4.410  &0.931  &4.74X \\\cline{1-5}
GoogLeNet-Inception5b.5&        0.95    &1.560  &0.220  &7.11X \\\cline{1-5}
\end{tabular}
\end{table}

\section{Conclusion and Future Work}
The high sparsity of output of ReLUs are explored in this paper and this property is exploited to accelerate the calculation of convolution layers by skipping zero-valued neurons. We first demostrate the sparsity of ReLUs on some popular deep convolutional networks, which displays that many ReLU layers have more than 70\% zero-valued outputs. And then we propose an efficient convolution algorithm by skipping the operations on zero-valued neurons, and it performs speedup improvement compared to the traditional convolution algorithm on our tested CPU platform.

Even though the algorithm proposed in this paper has some acceleration on CPUs, it is not comparable with the algorithms on GPUs. Therefore, in the future work, we need to design the sparse convolution algorithm on GPUs. 

\bibliographystyle{plain}
\bibliography{scnnv2.bbl}

\end{document}